\documentclass[journal]{IEEEtran}
\usepackage{float}

\usepackage{graphicx}
\usepackage{color}
\usepackage{amsmath}
\usepackage{threeparttable}
\newcommand\todo[1]{\textcolor{red}{#1}}
\usepackage{amssymb}
\usepackage[noadjust]{cite}
\usepackage{filecontents}

\usepackage{multicol}
\usepackage{fancyhdr}

\ifCLASSINFOpdf
\else
\fi
\begin{document}

%
\title{Multi-task Learning for Chest X-ray Abnormality Classification on Noisy Labels}

%
%

\author{Sebastian~G\"undel, 
        Florin~C.~Ghesu, Sasa~Grbic, Eli~Gibson, Bogdan~Georgescu, \textit{Member, IEEE,}\newline Andreas~Maier, \textit{Member, IEEE,} and Dorin~Comaniciu, \textit{Fellow, IEEE}
\thanks{S. G\"undel is with Siemens Healthineers, Medical Imaging
Technologies, Princeton, NJ 08540, USA and with the Pattern Recognition Lab, Friedrich-Alexander-Universit\"at Erlangen-N\"urnberg, 91058 Erlangen, Germany (e-mail: sebastian.guendel@fau.de)}
\thanks{F. Ghesu, S. Grbic, E. Gibson, B. Georgescu, and D. Comaniciu are with Siemens Healthineers, Medical Imaging Technologies, Princeton, NJ 08540, USA.}
\thanks{A. Maier is with the Pattern Recognition Lab, Friedrich-Alexander-Universit\"at Erlangen-N\"urnberg, 91058 Erlangen, Germany.}}

\maketitle

\begin{abstract}
Chest X-ray (CXR) is the most common X-ray examination performed in daily clinical practice for the diagnosis of various heart and lung abnormalities. The large amount of data to be read and reported, with 100+ studies per day for a single radiologist, poses a challenge in maintaining consistently high interpretation accuracy. In this work, we propose a method for the classification of different abnormalities based on CXR scans of the human body. The system is based on a novel multi-task deep learning architecture that in addition to the abnormality classification, supports the segmentation of the lungs and heart and classification of regions where the abnormality is located. We demonstrate that by training these tasks concurrently, one can increase the classification performance of the model. Experiments were performed on an extensive collection of 297,541 chest X-ray images from 86,876 patients, leading to a state-of-the-art performance level of 0.883 AUC on average for 12 different abnormalities. We also conducted a detailed performance analysis and compared the accuracy of our system with 3 board-certified radiologists. In this context, we highlight the high level of label noise inherent to this problem. On a reduced subset containing only cases with high confidence reference labels based on the consensus of the 3 radiologists, our system reached an average AUC of 0.945.

\end{abstract}



\begin{figure*}

\begin{center}
\includegraphics[width=7in]{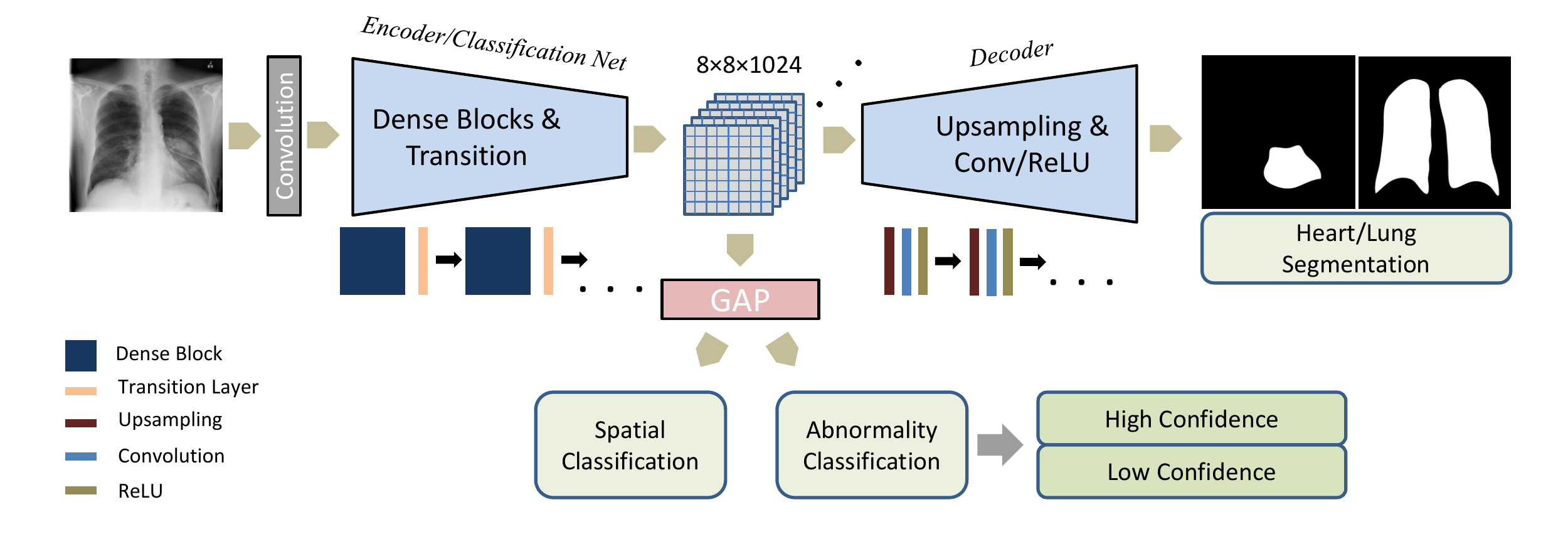}
\caption{Our proposed multi-task network predicting different abnormalities and approximate location (via classification). It consists of densely connected blocks followed by a global average pooling (GAP) layer to predict classification scores. Lung and heart masks are predicted with a connected decoder part.}
\label{fig:intro}
\end{center}
\end{figure*}

%

\section{Introduction}
\label{sec:intro}

\IEEEPARstart{R}{ecent} developments in the deep learning community combined with the availability of large annotated datasets have enabled the training of automated systems that can reach super-human performance on a variety of classification, detection and segmentation tasks \cite{Li2016PulmonaryNC, 8187667,Zhu2018DeepLungD3}. In different scenarios, such systems actively support humans, increasing the efficiency and accuracy of their workflow. In the medical domain, deep learning systems for image and data analysis and integration can potentially have an even greater impact, supporting the clinical workflow from patient admission to diagnosis, treatment and follow-up investigations \cite{De_Fauw_2018, Rajkomar_2018}.\\

In this paper, we focus on the problem of diagnosing multiple abnormalities based on chest radiographs (CXR) of the human body. In practice, this is a challenging problem reflected in a significant inter-user variability between different radiologists \cite{doi:10.1148/rg.2015150023}. The main reasons include the complex appearance of pathologies in X-ray projection images, and the large number of scans that need to be read and analyzed daily under time pressure \cite{article_error}. The average time to read and report a plain film is 1.4 minutes \cite{FLEISHON2006453}. \\

To address this challenge, we propose a method that can support the automatic classification of 12 abnormalities visible in chest X-rays. The system is based on a novel deep learning architecture which is able to predict - in addition to classification scores of abnormalities - lung/heart masks and classes about the location of certain abnormalities. By embedding the additional learning tasks, we observe a performance gain for abnormality classification. 

Since we encountered a high ratio of class label noise in the dataset, we conducted an observer study wherein two additional radiologists re-labeled a subset of the data. Based on the new annotations, strategies were applied to change the original dataset labels and to remove uncertain cases for a significant performance gain of our learned model. Furthermore, a strong correlation of most abnormalities between radiologist consensus and the remaining, certain cases can be seen.

\textbf{The contributions of this paper are as follows:}
\begin{itemize}
\item{We propose a novel multi-task deep neural network for multi-abnormality classification, spatial classification and lung/heart segmentation based on coronal chest X-ray images.}
\item{We demonstrate that the use of additional spatial knowledge related to pathologies or the underlying anatomical structures, i.e., heart and lungs, can significantly increase the accuracy of the abnormality classification.}
\item{To cope with the brightness and contrast variability of images from different sources, we propose a novel normalization technique. With this method we not only increase performance, but also significantly accelerate the training time.}
\item{We study the inter-user variability for this problem based on the input of three board-certified radiologists.}
\item{Based on the annotations of the three radiologists, we create a reduced evaluation subset using different voting strategies and show that the performance of our system increases.}
\item{Finally, we demonstrate the high correlation between the network classification probability and the multi-reader agreement.}
\end{itemize}

This paper builds on our preliminary work in X-ray abnormality detection systems \cite{DBLP:journals/corr/abs-1803-04565} with the following additional contributions: (a) a multi-radiologist observer study with extensive analysis of label noise and its impact on based on the detection system, (b) a task-specific normalization technique to increase robustness to variability caused by different acquisition equipment or post-processing steps, (c) an extension of the multi-task network for additional prediction of segmentation tasks, and (d) the prediction of confidence scores in addition to the classification of abnormalities.

%
%
%
%

\section{Related Work}
\label{sec:related}

The publication of the ChestX-ray14 (NIH) dataset  \cite{wang2017chestx}  has led to series of recent publications that propose automatic systems for abnormality classification. Wang et al. \cite{wang2017chestx} evaluated several state-of-the-art convolutional neural network architectures, reporting an area under the ROC curve (AUC) of 0.75 on average. Islam et al. \cite{DBLP:journals/corr/IslamAMA17} defined an ensemble of multiple state-of-the-art network architectures to increase the classification performance. Rajpurkar et al. \cite{rajpurkar2017chexnet} demonstrated that a common DenseNet architecture \cite{huang2018archive} can surpass the accuracy of radiologists in detecting pneumonia. In addition to a DenseNet, Yao et al. \cite{yao2017learning} implemented a Long-short Term Memory (LSTM) model to exploit dependencies between the abnormalities. An attention guided convolutional neural network architecture was used by Guan et al. \cite{2018arXiv180109927G} to specifically focus on the region of interest which is provided in a second network branch as a cropped image with higher resolution. Rubin et al. \cite{DBLP:journals/corr/abs-1804-07839} designed a Dual-Network to extract the image information of both frontal and lateral views. Yan et al. \cite{DBLP:journals/corr/abs-1807-06067} used a DenseNet architecture integrated with "Squeeze-and-Excitation'' blocks \cite{DBLP:journals/corr/abs-1709-01507} to improve the performance.

Wang et al. \cite{ChestNet} developed a new network architecture with both a classification and an attention branch, where the latter calculates activation maps with gradient-weighted class activation mapping \cite{DBLP:journals/corr/BaMK14} which is subsequently concatenated with the classification branch. Based on very limited location annotations of the abnormalities, Li et al. \cite{DBLP:journals/corr/abs-1711-06373} trained a neural network to predict both classification and localization of the abnormalities. Liu et al. \cite{liu2018sdfn} designed a network architecture with two branches, similar to \cite{2018arXiv180109927G}, where the second branch used a cropped image input based on existing lung masks. Yao et al. \cite{DBLP:journals/corr/abs-1803-07703} defined a network architecture which can be trained on different resolutions. Rajpurkar et al. \cite{10.1371/journal.pmed.1002686} used multiple radiologists to reannotate the images: One subgroup of radiologists defined the ground truth of a set where the other subgroup and the neural network was evaluated on. In this way, performance of both the radiologists and the deep learning algorithm could be compared. Irvin et al. \cite{irvin2019chexpert} trained on a dataset where the ground truth consists of an additional uncertainty class. Different approaches were applied during training to increase the performance with the uncertainty information.

We emphasize that most of the published works report classification results by splitting the data completely randomly for training, validation and testing \cite{yao2017learning,2018arXiv180109927G,DBLP:journals/corr/abs-1711-06373,DBLP:journals/corr/IslamAMA17}. With this splitting strategy, images of the same patient may be located in both training and testing set. For example the ChestX-ray14 dataset has an average of 3.6 images per patient. For a fair performance evaluation, the splitting should always be performed at patient level. The official split for the ChestX-ray14 data is performed patient wise. In our work, we use the official split. Moreover, for a valid performance comparison, same splits should be used since there is a significant performance variability by using different test sets \cite{DBLP:journals/corr/abs-1803-04565}.

\section{Problem Definition and Methodology}
\label{sec:Classification}

Given an arbitrary anterior-posterior (AP) or posterior-anterior (PA) chest X-ray image $\mathbf{I}$ with size $N\times N$, we design a deep learning based system parametrized by $\theta$ which outputs the probability of different abnormalities being present in the image: $\vec{o} = p(\mathbf{I}; \theta)$, where $\vec{o}\in[0,1]^D$ and $\mathbf{D}$ is the number of considered abnormalities. 
In addition, the system is designed to compute a probabilistic segmentation map $\mathbf{S}\in[0,1]^{2\times N\times N}$ for both lung lobes and the heart. 

\subsection{Dataset}

\begin{figure}[ht]
\begin{center}
\includegraphics[width=3.5in]{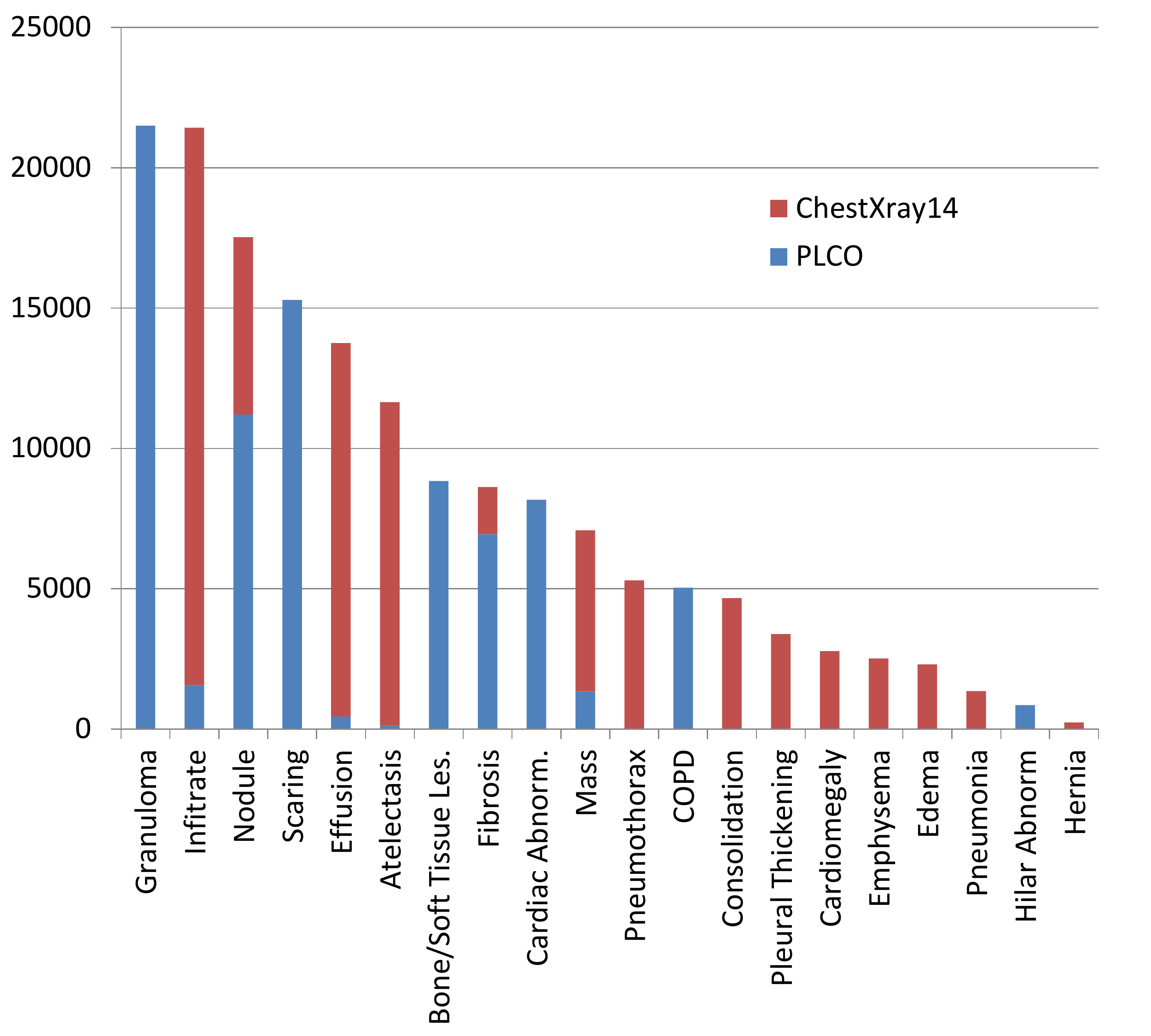}
\caption{This graph shows the number of images along with all abnormalities. The chart excludes the number of images where none of these pathologies appear.}
\label{fig:num_images}
\end{center}
\end{figure}

Our data collection is composed of two different datasets. The ChestX-Ray14 (NIH) \cite{wang2017chestx} and PLCO \cite{gohagan2000prostate} dataset. These datasets differ in several aspects. Table \ref{tab:dataset_overview} gives an overview. By combining both datasets, we can make use of 297,541 frontal chest X-ray images from 86,876 patients. Due to follow-up scans, there is an average of 3-4 images per patient. Therefore, patient-wise splits are considered for all experiments to separate the patients into training, validation, and test set. The PLCO dataset includes spatial information for some abnormality classes. Figure \ref{fig:num_images} shows the number of images observed to contain each abnormality. One image can also show multiple abnormalities. Additionally, the collections contain 178,319 images where none of the mentioned abnormalities appear, these images are not counted in Figure \ref{fig:num_images}.

\begin{table}[t]
\begin{center}
\begin{tabular}{| l || c | c |} 

\hline
Name & ChestX-Ray14 & PLCO\\
\hline

Number of images & 112,120 &  185,421 \\
Number of patients &  30,805 & 56,071 \\
Avg. image number per patient & 3.6 & 3.3 \\
Number of abnormalities & 14 & 12 \\
Image size & $1024\times1024$ & $\sim2500\times2100$ \\
Spatial information of abnorm. & no & partly \\
\hline

\end{tabular}
\\
\end{center}
\caption{Overview of the 2 datasets. Combining both, a new data collection with 297,541 images from 86,876 patients was created.}
\label{tab:dataset_overview}
\vspace{-0.15in}
\end{table}

Please note, the high imbalance of the data collection with respect to different abnormalities represents a challenge in ensuring training stability and performance.

\subsection{Deep Neural Network Design}
\label{subsec:NetworkDesign}
The classification branch of our multi-task network is inspired from the DenseNet architecture \cite{huang2018archive}. We adopt this network architecture with 5 dense blocks and a total of 121 convolutional layers (see Figure \ref{fig:intro} for an overview). Each dense block consist of several dense layers which include batch normalization, rectified linear units, and convolution. The novelty of the DenseNet are the skip connections, meaning that within a block, each layer is connected to all subsequent layers. Between each dense block, a so-called transition layer is added, which includes batch normalization, convolution, and pooling, to reduce the dimensions. \par 

The single grayscale input image $\mathbf{I}$ is rescaled to  $N\times N$ (in our experiments $N$=256 or $N$=512) using bilinear interpolation, replicated to 3 channels and fed into the network. The global average pooling (GAP) layer is resized depending on the input size. The number of output units is set to the number of abnormality classes $D$. We use sigmoid activation functions for each class to map the output to a probability interval $[0,1]$. The network is initialized with the pre-trained ImageNet model \cite{ImageNet}.

\subsection{Classification Training}
\label{subsec:Training}
A multi-label problem poses several challenges. The training process is modified such that each class can be trained individually. We create \textit{D} binary cross-entropy loss functions. The corresponding labels $[c_1, c_2 \ldots c_D] \in \{0,1\}$ (absence or presence of the abnormality, respectively) are compared with the network output and the loss is measured. Due to the highly imbalanced problem we introduce additional weight constants $w_P^{(n)}$ and $w_N^{(n)}$ to the cross-entropy function:
\begin{align}\label{eq:loss1}
\mathcal{L}_{Abn}^{(n)}(I,c_n) = -(&w_P^{(n)} * c_n\log(p_n) + \nonumber \\
&w_N^{(n)} * (1 - c_n)\log(1 - p_n)),
\end{align}                    

\begin{equation}\label{eq:loss1_g}
\begin{split}
\mathcal{L}_{Abn} = \sum_{n=1}^{D} \mathcal{L}_{Abn}^{(n)}(I,c_n),
\end{split}                    
\end{equation}
where $w_P^{(n)} = \frac{P_n + N_n}{P_n}$ and $w_N^{(n)} = \frac{P_n + N_n}{N_n}$, with $P_n$ and $N_n$ indicating the number of cases where the abnormality indexed by \textit{n} is present, respectively missing from the entire training set. For all experiments, we train with 128 samples in each batch. The Adam optimizer \cite{adam} ($\beta_1 = 0.9$, $\beta_2 = 0.999$, $\epsilon = 10^{-8}$)  is used with an adaptive learning rate: the learning rate is initialized with $10^{-3}$ and reduced by a factor of 10 when the validation loss plateaus. For the PLCO dataset, 70\% of the subjects were used for training, 10\% for validation, and 20\% for testing. For the ChestX-ray14 dataset, we use the provided ChestX-Ray14 benchmark split \cite{wang2017chestx}. \\

\textbf{Dataset Combination:} We propose to use two datasets that were acquired and annotated separately and differently. Both datasets contain several labels with the same definition as Figure \ref{fig:num_images} shows. A major issue of abnormality labeling is the varying and overlapping definition and interpretation between radiologists and abnormalities \cite{article_error}. Therefore, we treat the correspondent abnormalities of both datasets separately. Given $D_1=14$ abnormalities of the ChestX-ray14 dataset and $D_2=12$ abnormalities of the PLCO dataset, we define $D=D_1+D_2=26$ classes for our network. Furthermore, we only compute gradients for labels of one dataset where the current image is derived from. This strategy avoids a class categorization step beforehand and ensures that each network layer (except the last) receives information of all images. \\

\textbf{Normalization:} One challenge in processing chest radiographs is accounting for the large variability of the image appearance, depending on the acquisition source, radiation dose as well as proprietary non-linear postprocessing. In practice, one cannot systematically address this variation due to missing meta-information (e.g., unknown maximum high voltage for images from the NIH dataset). In this context, generic solutions have been proposed for the normalization of radiographs using multi-scale contrast enhancement/leveling techniques \cite{Philipsen2015LocalizedEN, 1000258}.

\begin{figure}[t]
\begin{center}
\includegraphics[width=3.5in]{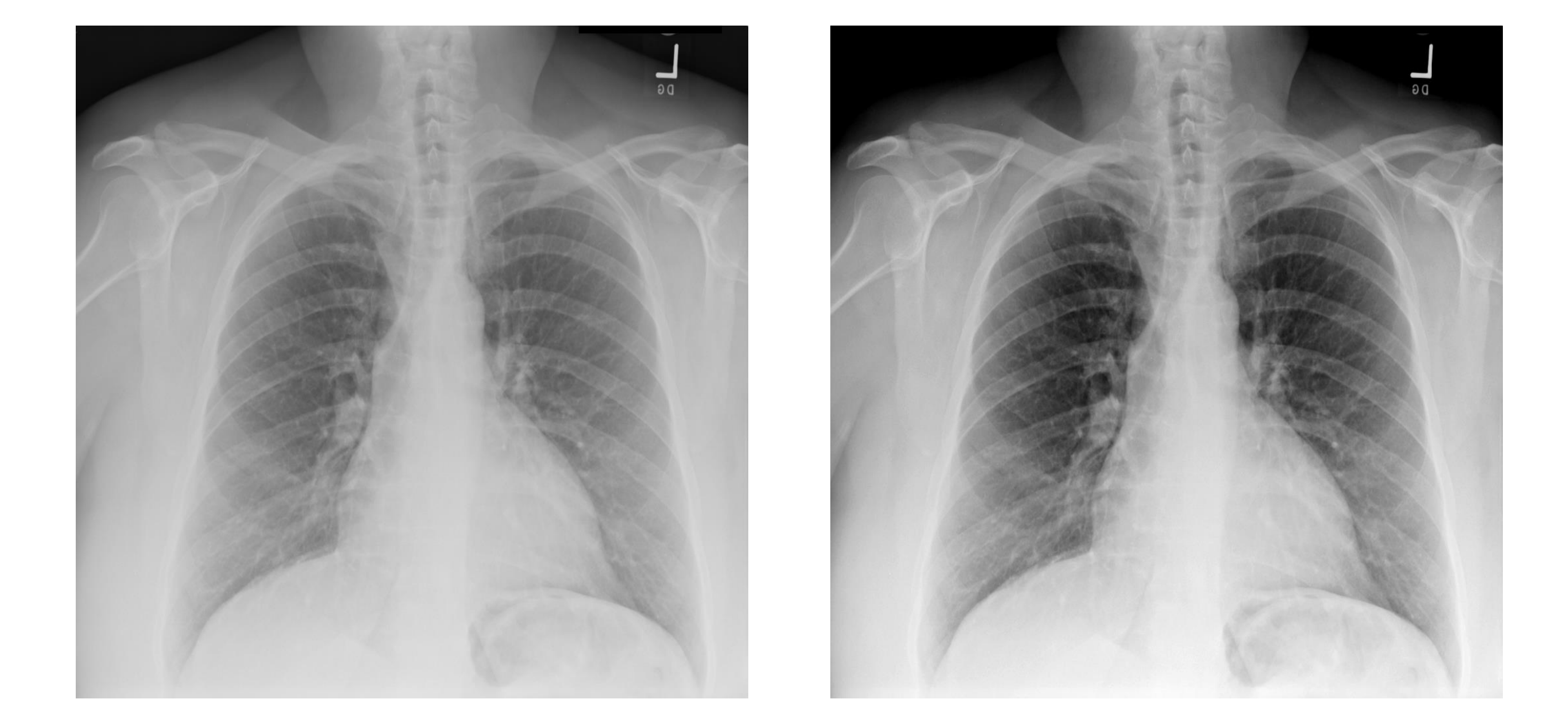}
\caption{An original image of the dataset is displayed (left) where the described normalization technique is applied (right).}
\label{fig:norm}
\end{center}
\end{figure}

For our diagnostic application, we propose to explicitly avoid altering the image appearance using one of these methods. Instead, we propose an efficient method for dynamically windowing each image, i.e., adjust the brightness and contrast via a linear transformation of the image intensities. Given an arbitrary chest X-ray image $\mathbf{I}$, let us denote its pixel value histogram function as $h(x;\mathbf{I})$. Using Gaussian smoothing and median filtering, one can significantly reduce the noise of $h$ (visible as, e.g., signal spikes due to black background or white text overlay) as well as account for long function tails that affect the windowing of the image. As such, based on the processed function $h$, we determine two bounds $b_{low}$, and $b_{high}$ which represent a tight intensity window for image I. The normalization is applied as follows, $\mathbf{I} = (\mathbf{I} - b_{low}) / (b_{high} - b_{low})$. A visual example is shown in Figure \ref{fig:norm}.

\section{Integrating Additional Knowledge}
\label{sec:additional_knowledge}

Additional spatial knowledge related to individual pathologies as well as the underlying anatomy, i.e., the heart and the lungs, can be exploited to increase the classification performance. Each individual feature described in this section lead to a moderate average performance gain. However, a combination of all features significantly improves the performance.

\subsection{Lung and Heart Segmentation}
\label{subsec:segmentation}
First, one can focus the learning task to the heart and lung regions. The image information outside of these regions may be regarded as irrelevant for the diagnosis of these lung/heart abnormalities.
\par

Instead of providing the predicted masks as input for the classification network, we extend the classification network with a decoder branch and predict the masks.
In this way, the additional knowledge about the shape of the heart and lung lobes is integrated in an implicit way, i.e., during learning through the flow of gradients. As such, in the encoder part, the network learns features that are not only relevant for the abnormality classification, but also for the isolation/segmentation of the relevant image regions.

\par

\begin{figure}[t]
\begin{center}
\includegraphics[width=3.5in]{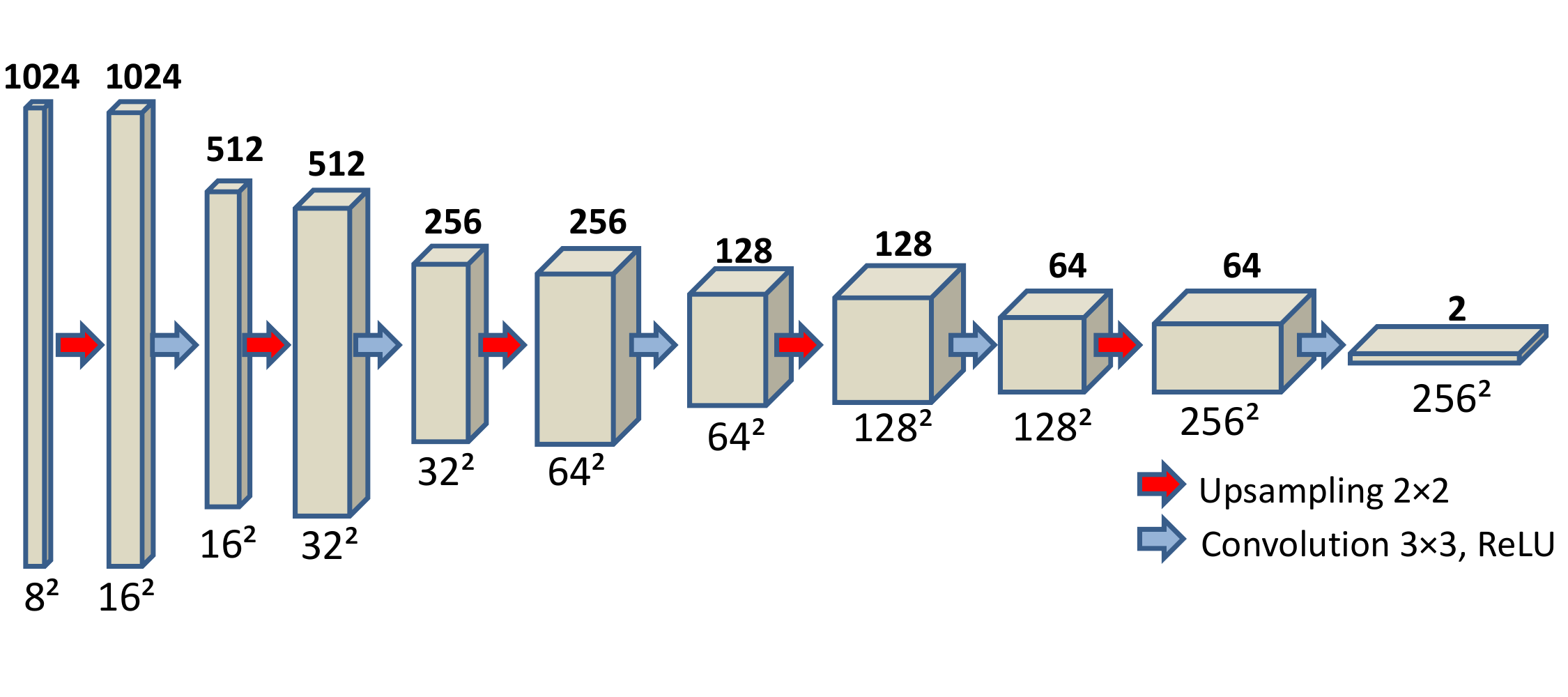}
\caption{Architecture of the decoder to predict segmentation masks. The network is connected to the classification network after the last dense block (left). The final layer predicts the lung and heart masks in 2 channels (right).}
\label{fig:decoder}
\end{center}
\end{figure}

The DenseNet model described in subsection \ref{subsec:NetworkDesign} is extended to solve the segmentation task. Therefore, we add a decoder network whose input is the returning feature maps of the last dense block. (see Figure \ref{fig:intro}). The decoder architecture is visualized in Figure \ref{fig:decoder}. For the segmentation task, we use the mean squared error loss function:

\begin{equation}\label{eq:mseloss}
\mathcal{L}_{Seg}(I,s) = \frac{1}{t}\sum\limits_{i=1}^t (s_{i} - p_{i})^2,
\end{equation}

where $t=2\times N\times N$ and $p_i\in \mathbf{S}$ denotes the output prediction of the current pixel $i$ and $s_i\in\{0,1\}$ the corresponding pixel label.

\subsection{Spatial Knowledge}

We propose to add additional supervision during learning using several approximate spatial labels provided with the PLCO data. For five abnormalities (Nodule, Mass, Infiltrate, Atelectasis, Hilar Abnormality) there is coarse location information available (see Table \ref{tab:region}). 

\begin{table}[ht]
\begin{center}
\begin{tabular}{| c | l || c | l |} 

\hline
No. & Region & No. & Region\\
\hline

1 & Left lobe  & 6 & Upper-middle part \\
2 & Right lobe & 7 & Upper part\\
3 & Lower part & 8 & Diffused (more attached parts)\\
4 & Lower-middle part & 9 & Multiple (more independent parts)\\
5 & Middle part & & \\

\hline

\end{tabular}
\\
\end{center}
\caption{Spatial Class Labels for the PLCO data}
\label{tab:region}
\vspace{-0.15in}
\end{table}

The location loss is another weighted cross-entropy loss with location-specific classes. The spatial labels $[b_1, b_2 \ldots b_F] \in \{0,1\}$, where \textit{F} is the total number of spatial classes listed in Table \ref{tab:region}, are compared with the network prediction and the loss is calculated (Equation \ref{eq:loss2} and \ref{eq:loss2_g}).

\begin{align}\label{eq:loss2}
\mathcal{L}_{Loc}^{(m)}(I,b_m) = -(&w_P^{(m)} * b_m\log(p_m) + \nonumber \\
&w_N^{(m)} * (1 - b_m)\log(1 - p_m)),
\end{align}

\begin{equation}\label{eq:loss2_g}
\begin{split}
\mathcal{L}_{Loc} = \sum_{n=1}^{F} \mathcal{L}_{Loc}^{(m)}(I,b_m),
\end{split}                    
\end{equation}

where $w_P^{(m)} = \frac{P_m + N_m}{P_m}$ and $w_N^{(m)} = \frac{P_m + N_m}{N_m}$, with $P_m$ and $N_m$ indicating, respectively, the number of presence and absence cases of spatial class \textit{m} in the training set. The individual localization loss
$\mathcal{L}_{Loc_m}$ is activated/deactivated dynamically: If spatial labels are not available for abnormality $n$, all spatial labels are disregarded and no gradients are computed. Otherwise, the loss is calculated as Equation \ref{eq:loss2} shows.

\textbf{Complete system training:} The global loss used for training is composed as follows:
\begin{equation}\label{eq:global_loss}
\begin{split}
\mathcal{L}_{Glob} = \mathcal{L}_{Abn} + \mathcal{L}_{Seg} + \mathcal{L}_{Loc}.
\end{split}                    
\end{equation}

The global architecture is shown in Figure \ref{fig:intro}.

\section{Experimental Results}
\label{sec:Experiments}

In our experiments, we measure the performance of our system at classifying different abnormalities, at segmenting the heart/lung region and at approximately localizing pathologies within the image. As a baseline, we measure test performance scores by training only the classification part on the PLCO dataset ($\mathcal{L}_{Seg} = \mathcal{L}_{Loc} = 0, \mathcal{L}_{Abn}\neq0$). The test set was evaluated with an AUC score of 0.859 (Table \ref{tab:classresults_all}, left).


\subsection{Lung and Heart Segmentation}

The second column in Table \ref{tab:classresults_all} shows improved classification scores when the network was additionally trained to generate lung and heart segmentation masks ($\mathcal{L}_{Seg} \neq0$). \par

\begin{figure}[ht]
\begin{center}
\includegraphics[width=3.5in]{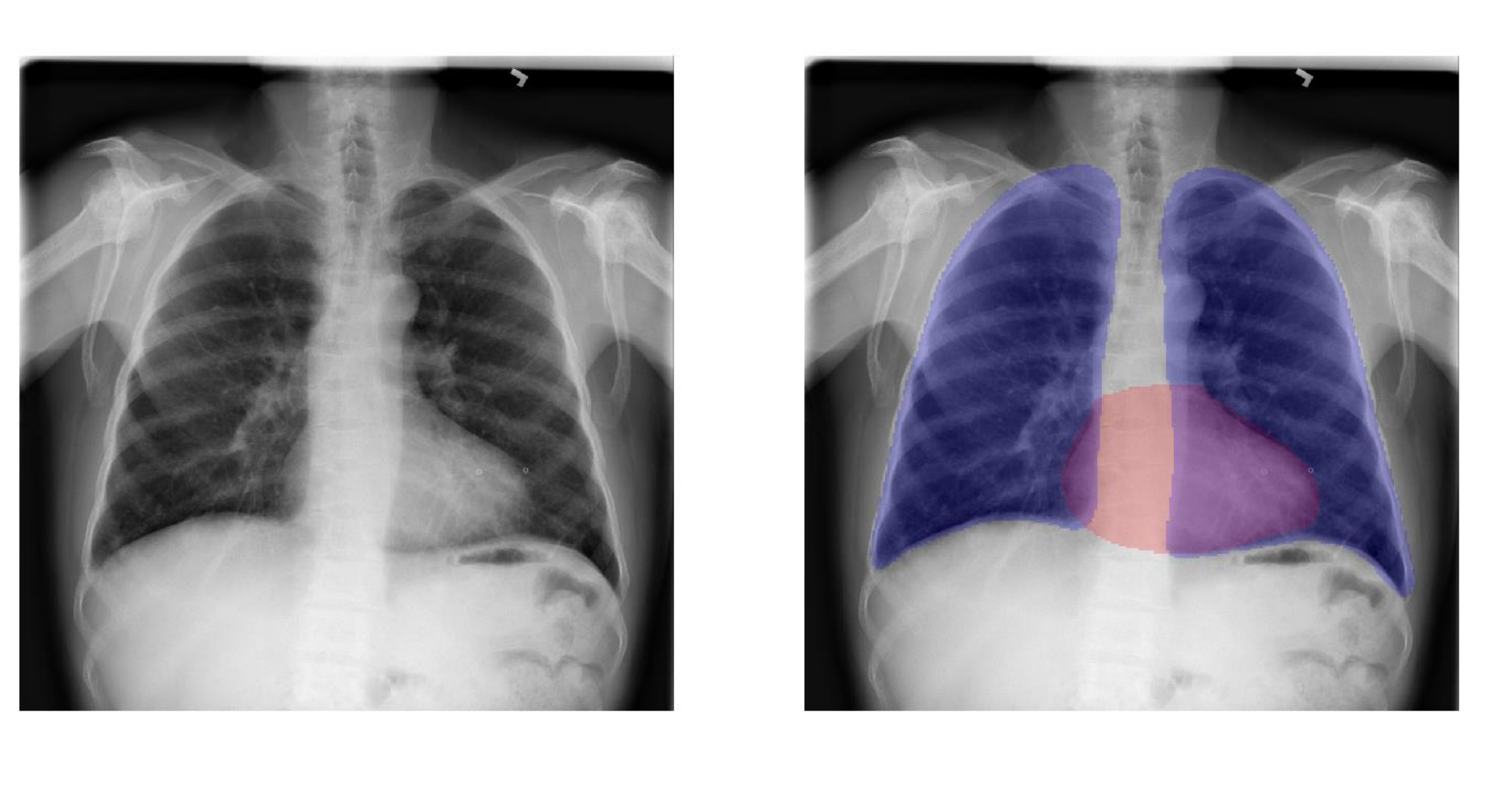}
\caption{Left: Example Chest X-ray image. Right: Predicted  segmentation masks of lung lobes (blue) and heart (red).}
\label{fig:segmentation}
\end{center}
\end{figure}

A test image is visualized in Figure \ref{fig:segmentation} (left). The probabilistic segmentation map is thresholded and overlayed with the image. The red mask defines the heart area, the blue mask indicates the two lungs in Figure \ref{fig:segmentation} (right). A quantitative evaluation of lung and heart segmentation was disregarded since we were focusing on the abnormality classification. The performance increased to 0.866 on average across the abnormalities (Table \ref{tab:classresults_all}, second Column).

\begin{figure}[ht]
\begin{center}
\includegraphics[width=3.5in]{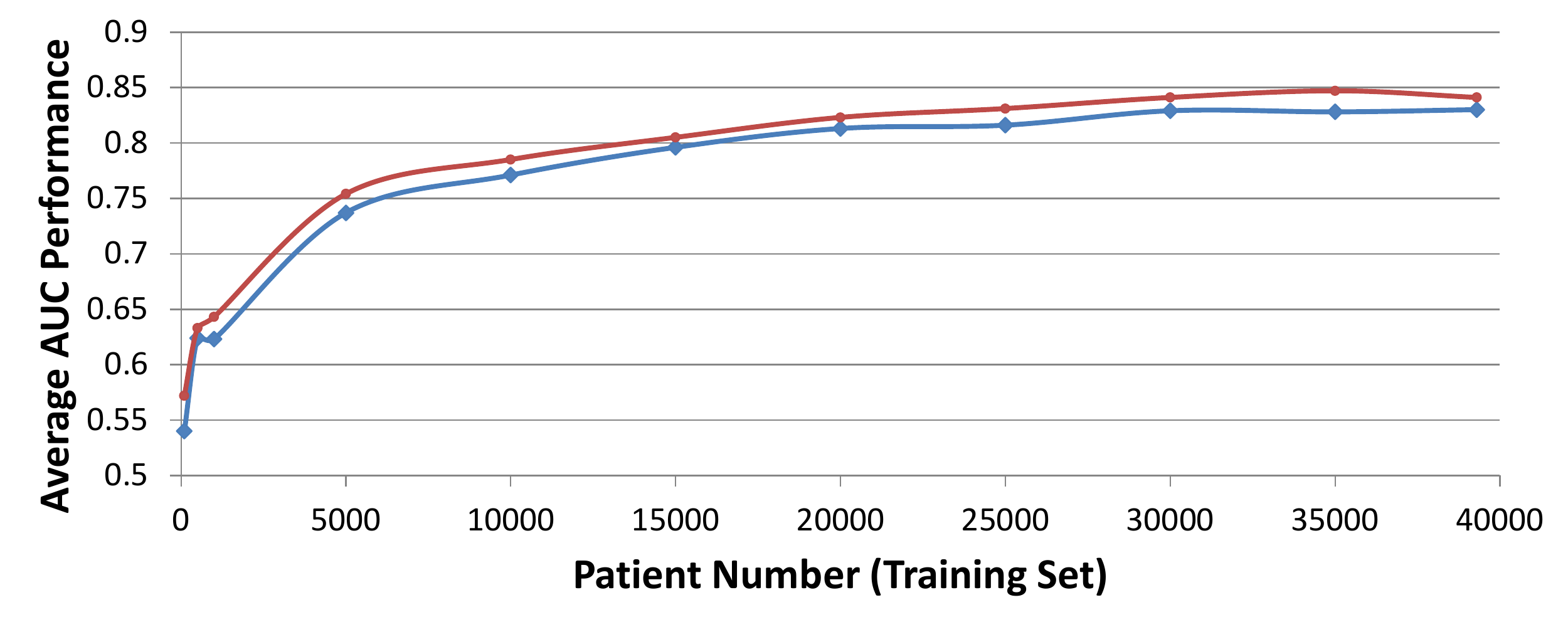}
\caption{Average AUC scores over the abnormalities with spatial information: Performance without (blue) and with (red) spatial information.}
\label{fig:location_performance}
\end{center}
\end{figure}

\begin{table*}[ht!]
\begin{center}
\begin{tabular}{| l || c | c | c | c | c | c |} 

\hline
Dim. Size & $256\times 256$ & $256\times 256$ & $256\times 256$ & $256\times 256$ & $256\times 256$ & $512\times 512$\\
\hline
Data & PLCO & PLCO & PLCO & PLCO+NIH & PLCO+NIH & PLCO+NIH\\
\hline
Features & - & Seg. & Loc. & Loc. & Loc.+Norm. & Loc.+Norm.+Seg\\
\hline
\hline
Nodule & 0.810 & 0.815 & 0.830 & 0.832 & 0.831 & 0.881\\
Mass & 0.829 & 0.839 & 0.840 & 0.867 & 0.869 & 0.884\\
Granuloma & 0.884 & 0.886 & 0.887 & 0.887 & 0.893 & 0.912\\
Infiltrate & 0.865 & 0.863 & 0.864 & 0.877 & 0.882 & 0.891\\
Scaring & 0.841 & 0.842 & 0.843 & 0.850 & 0.848 & 0.861\\
Fibrosis &  0.870 & 0.875 & 0.863 & 0.875 & 0.871 & 0.884\\
Bone/Soft Tissue Lesion & 0.841 & 0.846 & 0.834 & 0.840 & 0.850 & 0.848\\
Cardiac Abnormality & 0.926 & 0.928 & 0.922 & 0.923 & 0.927 & 0.926\\
COPD & 0.882 & 0.883 & 0.877 & 0.880 & 0.882 & 0.874\\
Effusion & 0.909 & 0.938 & 0.927 & 0.932 & 0.949 & 0.939\\
Atelectasis & 0.849 & 0.858 & 0.858 & 0.867 & 0.855 & 0.832\\
Hilar Abnormality & 0.796 & 0.815 & 0.812 & 0.815 & 0.850 & 0.859\\
\hline
Mean (Location Labels) & 0.830 & 0.838 & 0.841 & 0.852 & 0.857 & 0.869\\
\hline
\hline
Mean & 0.859 & 0.866 & 0.863 & 0.870 & 0.876 & \textbf{0.883}\\
\hline

\end{tabular}
\\
\end{center}
\caption{AUC classification scores for experiments tested on PLCO data}
\label{tab:classresults_all}
\vspace{-0.15in}
\end{table*}

\subsection{Spatial Knowledge}
\label{subsec:spatial_knowledge}
We measured the impact of the location labels on the performance of the classification. An average improvement of 0.011 (on abnormalities supported with spatial information) could be observed, as can be seen in the third column of Table \ref{tab:classresults_all}.

A more detailed experiment in Figure \ref{fig:location_performance} shows that increased patient numbers in the training set improved the test performance. The red curve indicates the average AUC performance on the abnormalities where both the spatial labels exist and localization classification during training is included. The blue one shows the average performance on the same abnormalities trained without classifying the spatial labels.

\subsection{All-in-One Joint Model}

By including the ChestX-ray 14 dataset, the average AUC score reaches 0.870. Performance values of each abnormality can be seen in Column 4 of Table \ref{tab:classresults_all}. Especially low frequency classes in the PLCO dataset were significantly improved, e.g., infiltrate (see Figure \ref{fig:num_images}) with 1,554 images by 0.013 when trained with 19.870 more images of the ChestX-ray 14 dataset.

\textbf{Normalization:} Including the normalization step based on dynamic windowing had a two-fold benefit. First, the training time was reduced on average 2-3 times. We hypothesize that this is because the normalization ensures that images are more aligned in terms of brightness and contrast, which in some sense simplifies the learning task. Second, this also improved the generalization of the model, and led to a performance increase to 0.876. \\

Finally, we upscaled the input image size to 512 in each dimension and changed the GAP layer of the DenseNet due to the bigger input size to $16\times16$. For the final network architecture we included the segmentation part. The average performance improved to 0.883 (see last column in Table \ref{tab:classresults_all}). Especially on small abnormalities, the classification performance improved significantly, e.g., nodules by 0.05. \\

With an average AUC performance of 0.883, our network achieves the best performance on such multi-abnormality problem in the community on a data collection trained and evaluated on the original dataset labels based on patient-wise train/test splits. \par

\begin{figure}[t]
\begin{center}
\vspace{-.1in}
\includegraphics[width=3.5in]{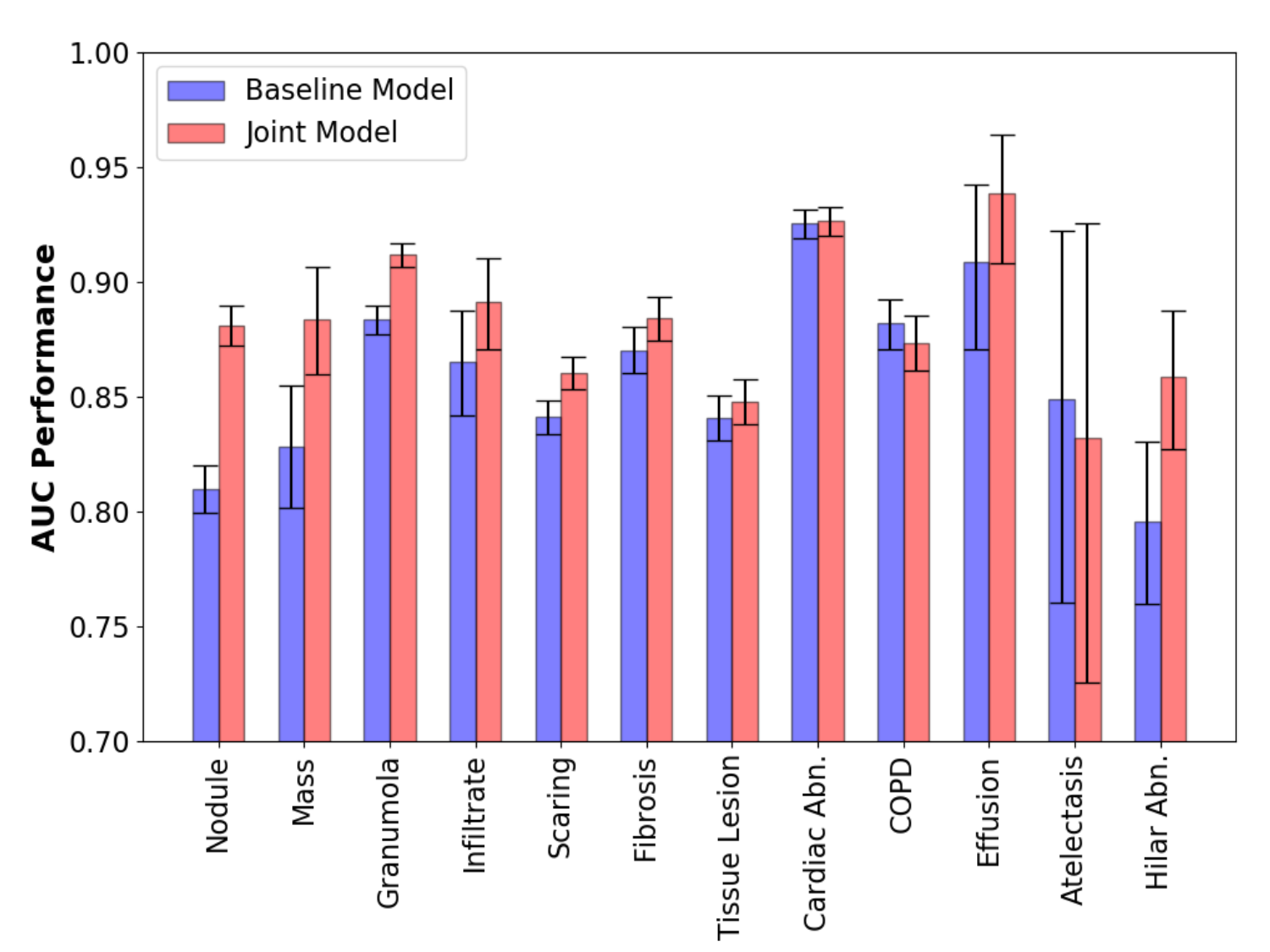}
\caption{AUC scores (bars) and 95\% confidence intervals using bootstrapping (whiskers) of all abnormalities for the baseline model and the joint model.}
\label{fig:bootstrapping}
\end{center}
\end{figure}

The integration of all described features lead to an average performance gain from 0.859 to 0.883. Some abnormalities could not benefit, e.g., cardiac abnormality, where the performance remained constant. However, other abnormalities significantly improved, e.g mass from 0.829 to 0.884. To show the statistical relevance, we applied bootstrapping on all abnormality classes. Figure \ref{fig:bootstrapping} shows the AUC scores of the baseline model (blue bars) and the joint model (red bars). The 95\% confidence intervals were added on each bar (black whiskers).

\subsection{Correlation between Accuracy and the Number of Patients in the Training Set}
\label{subsec:Pat_Depend}
\begin{figure}[t]
\begin{center}
\includegraphics[width=3.5in]{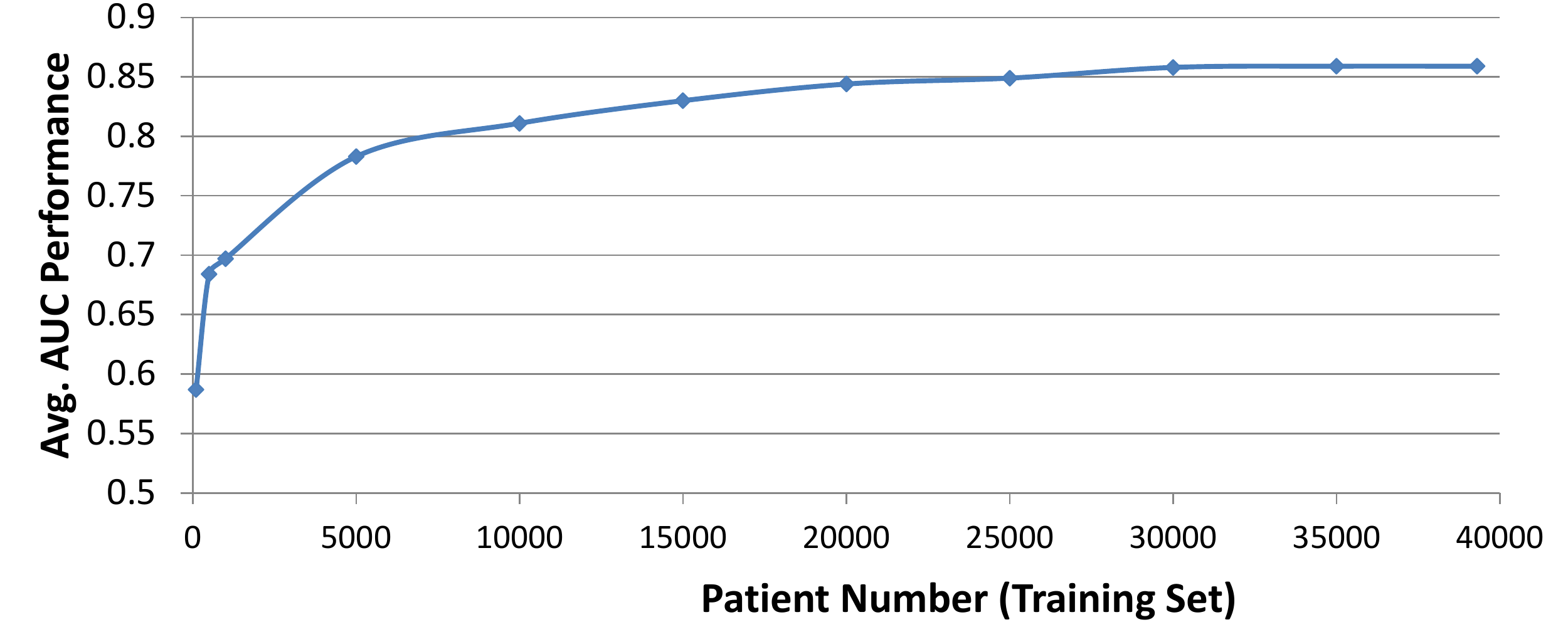}
\caption{AUC score over all abnormalities depending on the number of patients in the training set}
\label{fig:patient_number}
\end{center}
\end{figure}

A general question of any learning task is the required training set size to achieve state-of-the-art performance. In this experiment we focused on the PLCO dataset where a maximum of 70\% with 39,302 patients can be used in the training process. Several additional working points are processed after patients were randomly removed from the training set. We see the increasing average performance with increasing number of patients in the training set. However, the curve plateaus at a certain point (Figure \ref{fig:patient_number}).

\begin{figure}[t]
\begin{center}
\includegraphics[width=3.5in]{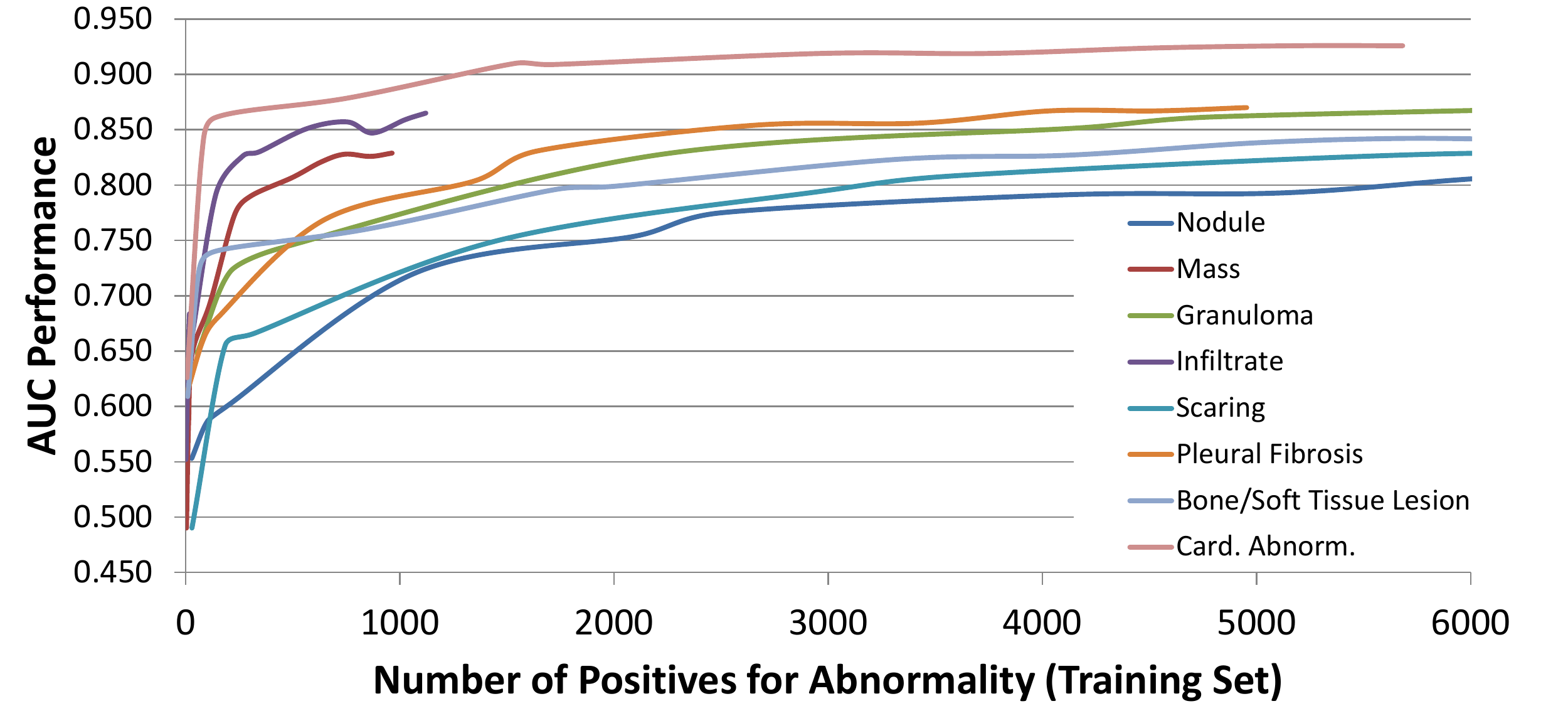}
\caption{AUC score of certain abnormalities depending on the number of images in the training set}
\label{fig:patient_number_abn}
\end{center}
\vspace{-0.2in}
\end{figure}

\begin{table*}[ht!]
\begin{center}
\begin{tabular}{| l | c c c | c c c c | c c c c c c |}

\hline
{} & \multicolumn{3}{c|}{\begin{tabular}{@{}c@{}} Complete \\ Agreement\end{tabular}} & \multicolumn{4}{c|}{Inter-rater Variability} & Rad. &\multicolumn{3}{c}{Majority Vote} & \multicolumn{1}{c}{\begin{tabular}{@{}c@{}}Complete \\Agreement\end{tabular}} & \multicolumn{1}{c|}{Band}\\
\hline
{} & Pos. & Neg. & No & PPA  & NPA & PD\textsubscript{PLCO} & $\kappa_F$ & AUC & Pos. & Neg. & AUC & AUC & AUC\\
\hline
Nodule & 11 & 464 & 90 & 0.307 & 0.964 & 0.480 & 0.313 & 0.743 & 31 & 534 & 0.845 & 0.951 & 0.911\\
Mass & 7 & 491 & 67 & 0.324 & 0.970 & 0.560 & 0.320 & 0.741 & 24 & 541 & 0.839 & 0.949 & 0.891\\
Granuloma & 4 & 503 & 58 & 0.242 & 0.980 & 0.740 & 0.248 & 0.835 & 15 & 550 & 0.945 & 1.000 & 0.975\\
Infiltrate & 7 & 447 & 111 & 0.254 & 0.959 & 0.560 & 0.212 & 0.765 & 30 & 535 & 0.850 & 0.898 & 0.909\\
Scaring & 5 & 422 & 138 & 0.364 & 0.916 & 0.580 & 0.218 & 0.745 & 52 & 513 & 0.816 & 0.876 & 0.892\\
Fibrosis & 15 & 419 & 131 & 0.349 & 0.935 & 0.340 & 0.294 & 0.752 & 51 & 514 & 0.860 & 0.959 & 0.918\\
Bone/Soft Tissue Lesion	& 22 & 326 & 217 & 0.343 & 0.890 & 0.080 & 0.207 & 0.864 & 82 & 483 & 0.794 & 0.949 & 0.815\\
Cardiac Abnormality & 17 & 397 & 151 & 0.435 & 0.898 & 0.240 & 0.310 & 0.918 & 73 & 492 & 0.927 & 0.963 & 0.947\\
COPD & 3 & 470 & 92 & 0.232 & 0.966 & 0.620 & 0.175 & 0.857 & 22 & 543 & 0.886 & 0.965 & 0.939\\
Effusion & 5 & 507 & 53 & 0.328 & 0.975 & 0.620 & 0.321 & 0.866 & 19 & 546 & 0.931 & 0.999 & 0.998\\
Atelectasis & 5 & 432 & 128 & 0.323 & 0.932 & 0.440 & 0.208 & 0.799 & 43 & 522 & 0.795 & 0.908 & 0.718\\
Hilar Abnormality & 4 & 473 & 88 & 0.239 & 0.968 & 0.660 & 0.198 & 0.718 & 22 & 543 & 0.842 & 0.924 & 0.893\\
\hline
Mean & {} & {} & {} & 0.312 & 0.946 & 0.493 & 0.252 & 0.800 & {} & {} & \textbf{0.861} & \textbf{0.945} & \textbf{0.901}\\
\hline

\end{tabular}
\\
\end{center}

\caption{Complete agreement of 3 readers - 2 Radiologists and the PLCO Labels (Left). Inter-rater variability measures of the three radiologists (Middle). Network performance evaluated on different ground truth: Single reader, majority vote, complete agreement and uncertainty band (Right).}
\label{tab:rad_eval_all}
\vspace{-0.15in}
\end{table*}

Focusing on single abnormalities, we gained valuable insight into the difficulty of the training process. Figure \ref{fig:patient_number_abn} shows performance scores of certain abnormalities, depending on the number of images. We can see that all abnormalities exhibit a similar curve structure; however, there is substantial variation between abnormalities in performance as the number of images increases. For a detailed analysis, we consider the two boundary classes: The cardiomegaly class reached the highest performance (top curve) at all working points. In contrast, the same amount of nodule images led to a substantially lower performance (bottom curve). We hypothesize that the performance difference can be explained based on the the abnormality characteristics: The variance of nodules is much higher, e.g., they can vary in shape, brightness, calcification, and other features \cite{nodule_variance}. Furthermore, they are much smaller and spread over the whole lung region \cite{GOULD2013e93S}. Cardiomegaly can always be recognized in the heart area \cite{cardiomegaly_size}. We also hypothesize that the difference is derived from abnormality dependent levels of label noise. A high amount of errors are caused by ``Satisfactory of Search'', meaning that a finding is missed because radiologists stop searching after finding the first abnormality \cite{foolmetwice}. \par 
In conclusion, a similar curve structure can be seen for all abnormalities. Thus, for low frequency abnormalities, e.g. mass and infiltrate, a performance gain can be expected when the training set contains more images including that abnormality. This hypothesis could be corroborated as we included the Chest-Xray14 dataset in Table \ref{tab:classresults_all}. The highest improvement by 0.027 can be seen on mass where the additional images are used.

\section{Multi-reader Validation}
Recent analysis \cite{ChestXray14Problems} has suggested that a large proportion of labels in the ChestX-ray14 dataset are inaccurate. Thus, we conducted an observer study with an additional two radiologists interpreting a subset of the data to measure (a) the performance of the system relative to that majority interpretation of multiple observers, and (b) to define a subset of cases where multiple radiologists agreed on the interpretation. \par

The experimental setup is originated from the publication by Batliner et al. \cite{univis90537614} where different voting strategies were investigated to increase the classification accuracy.

\subsection{Radiologist Agreement}

 Two board-certified radiologists with more then 15 years of experience in reading CXRs reannotated the data subset. Based on the PLCO data, we created a set containing 50 images per abnormality by random selection, avoiding more than one image per patient. Additionally, we included 50 images without any of those abnormalities. The new set, with a total of 565 images (resulting number of images due to a multi-label dataset) was read by the radiologists, deciding either on the presence or absence of all abnormalities based on the PLCO classes. We treated the PLCO labels as derived from one single reader. At first, we investigated the agreement of the 3 radiologists (2 radiologists + PLCO labels). In Table \ref{tab:rad_eval_all}, the positive, negative, and disagreement is listed (Column 1-3). A high number of cases without an agreement were observed.

\subsection{Inter-rater Reliability}
We evaluated statistical metrics to quantify consistency among the three ratings. First, we looked on the positive (PPA) and negative (NPA) predictive agreement which analyses the majority between the three radiologists. The values describe the ratio between the number of cases considered positive/negative by majority vote to the number of cases with a positive/negative finding from any of the radiologists. In Table \ref{tab:rad_eval_all} (Columns 4 and 5), values are listed for each abnormality. The low PPA values indicate a low agreement on individual findings. \\

With a positive disagreement (PD) check, we analyzed the variability between the original PLCO labels and our two radiologists (Column 6). PD\textsubscript{PLCO} is indicated as the number of positive PLCO cases provided that both radiologists disagreed with PLCO, divided by the total number of positive PLCO cases. Given that we included the same number of positive PLCO cases for all abnormalities, the PLCO reader labeled 49.3\% of the abnormalities as positive where our both radiologist labeled no finding. \\

The Fleiss' kappa value \cite{fleiss} is commonly used to measure the agreement between multiple raters. According to the kappa scale by Landis et al. \cite{Landis1977TheMO}, we have a fair agreement on the average kappa score of 0.252 (Column 7). \par

All observed parameters point out that individual radiologists include a strong bias when reading CXRs. A weak reliability could be observed, especially, when a reader reported an abnormality finding.

\subsection{Network Performance on refined Ground Truth }

For the multi-reader evaluation, the annotated subset is used as test set. The rest of the data is split into training (90\%) and validation (10\%). Overlapping subjects between the test and training/validation set were removed. Subsequently, the model was retrained. The evaluation can be seen in Table \ref{tab:rad_eval_all}, where the network performance is listed based on the PLCO label ground truth (Column 8 of Table \ref{tab:rad_eval_all}).

Due to the strong variability between readers, we defined our new ground truth as a majority vote of the three radiologists. Performance scores and the distribution of positive (Pos.) and negative (Neg.) cases are listed in Columns 9-11. The significant average performance gain to 0.861 supports the hypothesis that individual annotators have integrated a high label bias. Moreover, the model which is trained on biased labels (PLCO only), seems to be robust against this label noise. 

As a further step, we eliminated cases without a complete agreement between the three radiologists. In Table \ref{tab:rad_eval_all}, the test performance of the subset is listed (Column 12). On average, we reached a performance of 0.945.

With 50 positive PLCO cases per abnormality and the resulting low number of a complete agreement e.g. COPD (50 to 3), we speculate that the strong label bias in the PLCO dataset were derived from the usage of prior knowledge.\\

\begin{figure}[t]
\begin{center}
\includegraphics[width=3.5in]{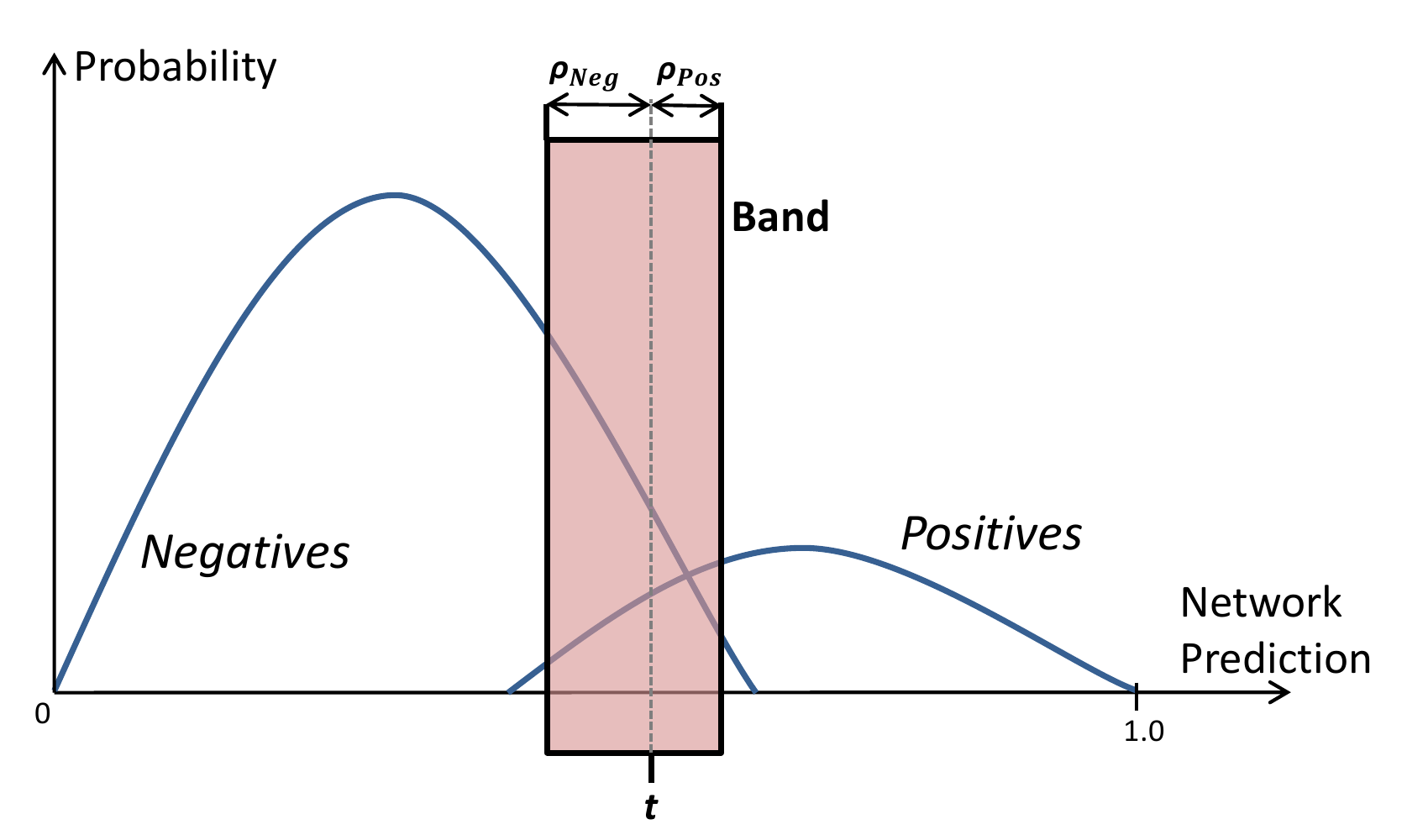}
\caption{Removal of cases within the band. The parameter $t$ is defined as the threshold of positive and negative predicted cases and parametrized to have the same number of false positive and false negative cases. The width $\rho_{Neg}$ and $\rho_{Pos}$ are set such that the band contains $l_{TP}$\% of true positives and $l_{TN}$\% of true negatives.}
\label{fig:band}
\end{center}
\end{figure}

\textbf{User Agreement versus Algorithm Confidence}: Under a hypothesis that radiologist agreement might correlate with the confidence of the algorithm, we investigated whether a band of algorithm probabilities could be defined from the validation set that would separate confidently true negatives, uncertain algorithm findings, and confidently true positives on unseen data.\\

Based on multiple radiologists, we defined confidences for each class in each image. Four different confidence categories were created: 
\begin{multicols}{2}
\begin{itemize}
    
    \item{High negative}
    \item{Low negative}
    \item{High positive}
    \item{Low positive}
    
\end{itemize}
\end{multicols}
 The confidence value for all classes and images depends on the number of positive label annotations of the three radiologists. An abnormality class classified as low positive confidence is derived from two readers reporting an abnormality where one reader reported no finding.\\

The AUC score is based on a threshold which is shifted over the positive and negative distribution. Dealing with high and low confident cases we defined an additional band over the output prediction range. Low confident cases may be located within the band and be misclassified. All samples included in the band were eliminated and, subsequently, the performance is measured on the reduced set. By including the band, we achieved the following contributions:
\begin{itemize}
    \item We significantly improved the AUC performance after removing cases within the band.
    \item We observed that for most abnormalities the majority of cases included within the band were low confidence cases.\\
\end{itemize} 

\begin{table}[t!]
\begin{center}
\begin{tabular}{| l | c | c | c | c |}

\hline
{} & \multicolumn{2}{c|}{Negative} & \multicolumn{2}{c|}{Positive} \\
\hline
{} & High Conf. & Low Conf. & Low Conf. & High Conf. \\

{} & 0/3 Positive & 1/3 Positive & 2/3 Positive & 3/3 Positive \\
\hline
{} & B / A / \% & B / A / \% & B / A / \% & B / A / \% \\
Nod. & 464 / 306 / \textbf{66} & 70 / 32 / \textbf{46} & 20 / 7 / \textbf{35} & 11 / 8 / \textbf{72}\\
Mass & 491 / 307 / \textbf{63} & 50 / 18 / \textbf{36} & 17 / 5 / 29 & 7 / 2 / 29\\
Gran. & 503 / 365 / \textbf{73} & 47 / 27 / \textbf{57} & 11 / 9 / \textbf{82} & 4 / 4 / \textbf{100}\\
Inf. & 447 / 297 / \textbf{66} & 88 / 38 / \textbf{43} & 23 / 8 / \textbf{35} & 7 / 3 / \textbf{42}\\
Scar. & 422 / 274 / \textbf{65} & 91 / 49 / \textbf{54} & 47 / 23 / \textbf{49} & 5 / 4 / \textbf{80}\\
Fibr. & 419 / 318 / \textbf{76} & 95 / 35 / \textbf{37} & 36 / 14 / \textbf{39} & 15 / 11 / \textbf{73}\\
Les. & 326 / 265 / \textbf{81} & 157 / 114 / \textbf{72} & 60 / 38 / \textbf{63} & 22 / 17 / \textbf{77}\\
Card. & 397 / 334 / \textbf{84} & 95 / 53 / \textbf{56} & 56 / 20 / \textbf{36} & 17 / 9 / \textbf{53}\\
CO. & 470 / 353 / \textbf{75} & 73 / 32 / \textbf{44} & 19 / 9 / 47 & 3 / 1 / 33\\
Eff. & 507 / 303 / \textbf{60} & 39 / 6 / \textbf{15} & 14 / 1 / \textbf{7} & 5 / 4 / \textbf{80}\\
Atel. & 432 / 284 / \textbf{66} & 90 / 31 / \textbf{34} & 38 / 7 / \textbf{18} & 5 / 1 / \textbf{20}\\
Hil. & 473 / 312 / \textbf{66} & 70 / 32 / \textbf{46} & 18 / 7 / 39 & 4 / 1 / 25\\
\hline
Avg. & 446 / 310 / \textbf{70} & 80 / 39 / \textbf{49} & 30 / 12 / \textbf{40} & 9 / 5
/ \textbf{56}\\
\hline

\end{tabular}
\\
\end{center}
\caption{Absolute values of cases Before (B) and After (A) the removal with the uncertainty band based on the defined confidence classes. The last value shows the percentage (\%) of cases maintained after band removal.}
\label{tab:band}
\end{table}

The band was defined as an interval $[t_n-\rho_{Neg}^{(n)},t_n+\rho_{Pos}^{(n)}]$. The parameters $t_n$ indicate the threshold of positive and negative cases for abnormality $n$ and the sum of $\rho_{Neg}^{(n)}$ and $\rho_{Pos}^{(n)}$ the width. The threshold is set such that we have equal false positive and false negative ratios for each abnormality. We allowed a maximum of $l_{TP}$\% of true positives and $l_{TN}$\% of true negatives to be inside the band. The values $\rho_{Neg}$ and $\rho_{Pos}$ are parametrized such that the condition is fulfilled. The parameters were found on the validation set, the evaluation is on the test set based on majority vote of the 3 readers. For our experiment, we used $l_{TP}$ = $l_{TN}$ = 20. An example scenario is illustrated in Figure \ref{fig:band}. The band contains almost all false positive and false negative cases. \par 

After elimination of cases within the band, the performance values are shown in Table \ref{tab:rad_eval_all} (last column) reaching 0.901 on average. Thus, a significant performance gain could be achieved when disregarding these cases. \\

After removing the cases within the band, the number of eliminated/maintained cases were analyzed with respect to the four defined confidence classes. Table \ref{tab:band} shows the absolute numbers before (B) and after (A) elimination of cases within the band. The last value indicates the percentage (\%) of each abnormality maintained after reducing the set. Most of the abnormalities show that significantly more high confident images were maintained after removal, e.g. fibrosis: 76\% of the high confident negative cases and 37\% of the low confident negative cases as well 73\% of high confident positive cases and 39\% of low confident positive cases could be maintained.

\section{Discussion}

We observed that our multi-task convolutional neural network is able to correctly classify a wide range of abnormalities in chest X-ray images. During the learning process of abnormality classification, the system was supported by additional features, e.g., spatial information and normalization. Due to the classification of multiple abnormalities with different characteristics, we found features which improved the AUC performance for most abnormalities. The combination of all integrated features lead to a significant average performance gain.

The low agreement of multiple readers disclosed the high bias between radiologists. Less confidence should be given to any individual dataset labels in future. If available, a combination of more radiologists should be always considered. A majority vote of three radiologists showed significantly improved results. We hypothesize that even more radiologists are necessary for a more trustful and unbiased evaluation. 

An evaluation of cases with a complete agreement between the radiologists additionally improved the performance. This improvement may be caused due to the elimination of challenging examples. With the help of the uncertainty band, a correlation between network prediction and radiologist agreement could be determined. Including this feature, a flag (in addition to presence/absence of an abnormality) could be returned by the system whether a certain case is band in- or outlying. The reading time of cases outside the band and, hence, high confident cases, could be reduced and used for cases with low confidence.

\section{Conclusion}

Overall, we developed a multi-task network for the classification of 12 different abnormalities, classification of their location, and segmentation of lung lobes and heart. The system was trained and evaluated on a data collection of 297,541 images. With additional information of lung and heart segmentations and spatial labels, as well as by using an adaptive normalization strategy, we significantly improved the abnormality classification performance to an average AUC score of 0.883 on 12 different abnormalities. An agreement study between the original labels and additional radiologist annotations indicated a large fraction of label noise. Re-labeling the original data annotations based on multi-reader voting showed a significant performance gain. By additionally removing uncertain cases we reached an average AUC performance of 0.945. Furthermore, we demonstrated that the derived confidence scores of most annotations are highly correlated with the prediction of the network. An uncertainty band integrated over the prediction range of the model filters out low confident cases for 

\section*{Acknowledgement}
The authors thank the National Cancer Institute (NCI) for access to their data collected by the Prostate, Lung, Colorectal and Ovarian (PLCO) Cancer Screening Trial. The authors thank the National Institutes of Health (NIH) for access to the ChestX-ray14 collection. The statements contained herein are solely those of the authors and do not represent or imply concurrence or endorsement by NCI or NIH.\newline
\textbf{Disclaimer: }The concepts and information presented in this paper are based on research results that are not commercially available.

\ifCLASSOPTIONcaptionsoff
  \newpage
\fi



\bibliographystyle{bibtex/IEEEtran}
%


\bibliography{ChestXrayJournal.bbl}


%

\end{document}